\documentclass{article}
\usepackage[utf8]{inputenc}
\usepackage{a4wide}
\usepackage{xcolor}
\usepackage{amsmath}
\usepackage{lscape}
\usepackage[font=small,labelfont=bf]{caption}
\usepackage{graphicx}
\usepackage{subcaption}
\usepackage{comment}
\usepackage{tabularx}
\usepackage{multirow}
\usepackage{algorithmic}
\usepackage[ruled,vlined]{algorithm2e}
\usepackage{amsmath,amssymb,amsfonts,amsthm} 
{

\theoremstyle{remark}

}
\usepackage[utf8]{inputenc}
\usepackage{amssymb}
\usepackage{multicol}
\usepackage{geometry}
\usepackage{hyperref}
\usepackage{authblk}
\usepackage{ulem}
\usepackage{moreverb}
\usepackage{amsmath,bm}
\usepackage{graphicx}
\usepackage{siunitx}
\usepackage{subcaption}

\renewenvironment{abstract}
  {\small\noindent\textbf{Abstract.}\normalsize\ignorespaces}
  {\par\noindent\ignorespacesafterend}

\makeatletter
\renewcommand{\maketitle}{%
    \noindent\fcolorbox{red}{white}{% Red border, white background
        \parbox{\textwidth}{%
            \color{red}Accepted for the \textit{15th International Conference on Swarm Intelligence (ANTS 2026)}. 
        }%
    }
    \vspace{1em} % Add some vertical space between the box and the title
    \begin{center}%
        {\LARGE \@title \par}% Title
        \vspace{0.5em}% Add some vertical space
        {\large \@author \par}% Author
        \vspace{0.5em}% Add some vertical space
        {\large \@date}% Date
    \end{center}%
    \vspace{1em} % Add some vertical space after the title
}
\makeatother

\begin{document}

\title{FDA Flocking: Future Direction-Aware Flocking via Velocity Prediction}

\author{Hossein B. Jond}
\author{Martin Saska}
\affil{Department of Cybernetics, Czech Technical University in Prague, Prague, Czechia}

\date{}

\maketitle

\begin{abstract}
Understanding self-organization in natural collectives such as bird flocks inspires swarm robotics, yet most flocking models remain reactive, overlooking anticipatory cues that enhance coordination. Motivated by avian postural and wingbeat signals, as well as multirotor attitude tilts that precede directional changes, this work introduces a principled, bio-inspired anticipatory augmentation of reactive flocking termed \textit{Future Direction-Aware} (FDA) flocking. In the proposed framework, agents blend reactive alignment with a predictive term based on short-term estimates of neighbors’ future velocities, regulated by a tunable blending parameter that interpolates between reactive and anticipatory behaviors. This predictive structure enhances velocity consensus and cohesion–separation balance while mitigating the adverse effects of sensing and communication delays and measurement noise that destabilize reactive baselines. Simulation results demonstrate that FDA achieves faster and higher alignment, enhanced translational displacement of the flock, and improved robustness to delays and noise compared to a purely reactive model. Future work will investigate adaptive blending strategies, weighted prediction schemes, and experimental validation on multirotor drone swarms.
\end{abstract}

\section{Introduction}\label{sec:introduction}

The self-organization of large groups of agents into coherent spatiotemporal patterns—such as bird flocks and fish schools—emerges from local perception-based interactions that produce collective motion without centralized control. Reynolds' Boids model \cite{reynolds1987flocks} captured this through three reactive rules: cohesion (proximity maintenance), separation (collision avoidance), and alignment (velocity matching). Subsequent influential models, including Vicsek \cite{vicsek1995novel}, Cucker-Smale \cite{cucker2007emergent}, Couzin's zonal scheme \cite{couzin2002collective}, and various multi-agent robotic swarms \cite{olfati2006flocking,viragh2014flocking,han2024multi,jond2025bearing}, largely retain these reactive principles.

However, purely reactive approaches, which respond only to instantaneous neighbor states, struggle to replicate the smooth, anticipatory coordination observed in biological flocks. Empirical studies show that birds use subtle postural adjustments, wing orientations, and wingbeat signals to predict neighbors' impending maneuvers, enabling coherent propagation of turns and minimizing damping of directional information \cite{cavagna2015flocking,cavagna2018physics,ballerini2008empirical,cavagna2010scale,castro2024modeling}. Reactive models (e.g., Vicsek) dissipate such predictive cues, leading to less robust collective turns and reduced performance under realistic conditions. This biological insight motivates augmenting reactive rules with anticipation, as explored in recent visual servoing and robotic swarm research using optic flow or learned features \cite{moshtagh2009vision,schilling2021vision,mezey2025purely}.

\begin{figure}[t]
\centering
    \begin{minipage}{0.45\textwidth}
        \centering
        \includegraphics[width=\linewidth]{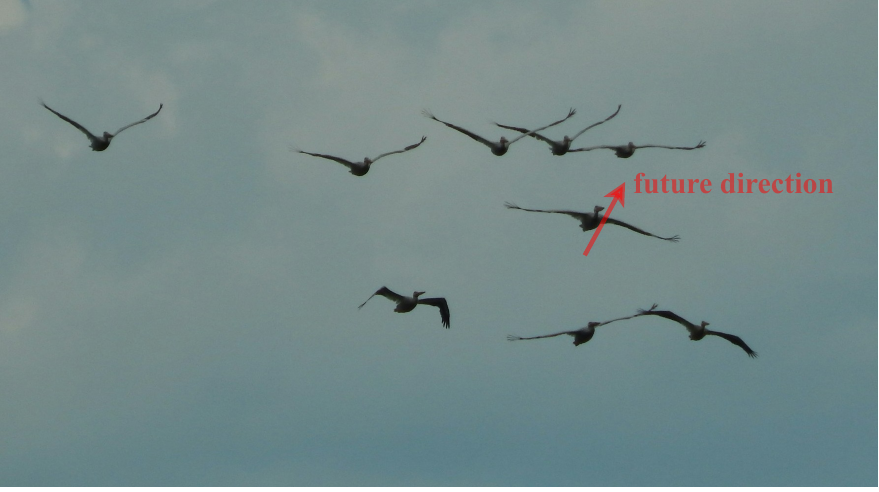}
        {\small (a)}
    \end{minipage}\hfill
    \begin{minipage}{0.45\textwidth}
        \centering
        \includegraphics[width=\linewidth]{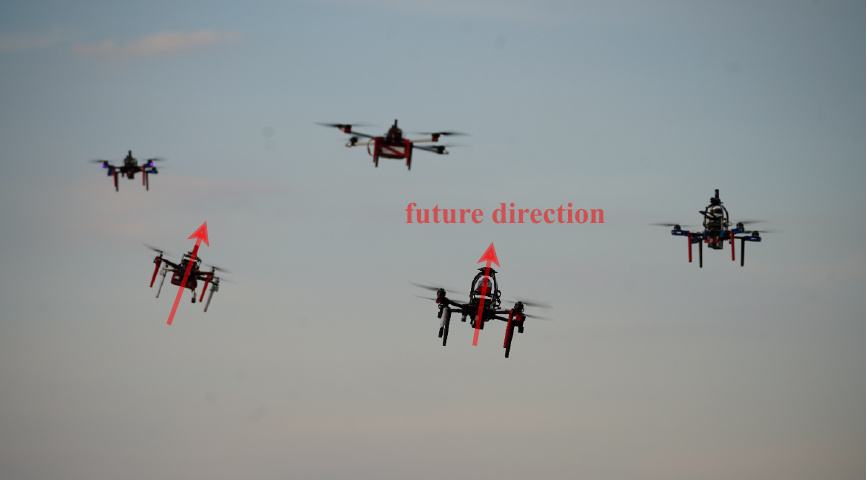}
        {\small (b)}
    \end{minipage}
\caption{Anticipatory cues in natural and engineered collectives: (a) In bird flocks, individuals signal impending maneuvers via body posture and wing alignment, which precede direction changes. Neighbors interpret these cues to predict their future direction, ensuring smooth coordination; (b) In multirotor drones, attitude tilting (pitch and roll) acts as an analog, as drones must tilt before accelerating in a new direction, providing an implicit cue of future velocity.}
\label{fig:anticipatory-cues}
\end{figure}

We provide a principled, bio-inspired anticipatory augmentation of reactive flocking—termed Future Direction-Aware (FDA) flocking—and characterize its performance under sensing and communication imperfections. Drawing from natural flocks, where postural and wingbeat cues forecast future velocities \cite{hoetzlein2024flock2}, and from multirotor drones, where attitude tilting (pitch/roll) signals intended acceleration \cite{vasarhelyi2018optimized}, FDA agents blend reactive alignment with a predictive term based on short-term neighbor velocity estimates. A tunable blending parameter balances immediate feedback against forward-looking adjustments, yielding improved velocity consensus, cohesion–separation balance, and collision avoidance, while mitigating the adverse effects of perception/communication delays and sensor noise that destabilize purely reactive models \cite{wang2019impacts,munz2008delay,shi2025flocking,vasarhelyi2018optimized,casas2018impact}. Fig.~\ref{fig:anticipatory-cues} illustrates these anticipatory cues in birds and drones, highlighting their role in enabling predictive coordination.

Simulations demonstrate that the FDA achieves faster and higher alignment, steadier centroid trajectories, and substantially less performance degradation under perturbations compared to the reactive baseline. While the current model uses uniform averaging of predictions (potentially missing nuanced weighting of stronger cues as in biology), future extensions will include adaptive blending, weighted prediction schemes, and real-world robotic validation.

The remainder of this paper is organized as follows. Section~\ref{sec:reactive_flocking} reviews the baseline reactive flocking model. Section~\ref{sec:fda_flocking} presents the FDA framework, velocity prediction, and extensions for delay and noise. Section~\ref{sec:simulation_results} reports simulation results on performance and robustness. Section~\ref{sec:conclusion} concludes with key insights and future directions.

\section{Reactive Flocking}\label{sec:reactive_flocking}

Consider \(n \geq 2\) agents indexed by \(\mathcal{N} = \{1, \dots, n\}\) in \(\mathbb{R}^m\). Each agent \(i \in \mathcal{N}\) has position \(\mathbf{p}_i\), velocity \(\mathbf{v}_i\), and acceleration/control \(\mathbf{u}_i\) in \(\mathbb{R}^m\). Its neighborhood is \(\mathcal{N}_i = \{j \neq i \mid \|\mathbf{p}_j - \mathbf{p}_i\| \leq r_i\}\), with interaction radius \(r_i > 0\).

The flocking model is based on \cite{jond2025minimal} and is given by
\begin{align}\label{eq:flock-model}
\begin{cases} 
    \dot{\mathbf{p}}_i &= \mathbf{v}_i, \quad \dot{\mathbf{v}}_i = \mathbf{u}_i, \\
    \mathbf{u}_i &=\sum_{j \in \mathcal{N}_i} \psi(\|\mathbf{p}_j - \mathbf{p}_i\|)(\mathbf{p}_j - \mathbf{p}_i) +\phi_i \sum_{j \in \mathcal{N}_i}  (\mathbf{v}_j - \mathbf{v}_i),
    \end{cases}
\end{align}
where \(\psi(\|\mathbf{p}_j - \mathbf{p}_i\|) = 1 - \frac{\delta_i|\mathcal{N}_i|}{\|\mathbf{p}_j - \mathbf{p}_i\|}\), \(\|\mathbf{p}_j - \mathbf{p}_i\|>0\), \(\phi_i = \frac{1}{|\mathcal{N}_i|}\), \(\mathcal{N}_i \neq \emptyset\), are the interaction weights, \(\delta_i\geq0\) is the spatial offset, and \(|\mathcal{N}_i|\) is the neighborhood size. Velocity and control inputs are obtained via smooth saturation as
\begin{equation}  \label{eq:saturation}
\mathbf{x}_i = x_i^{\max} \tanh\left(\frac{\|\mathbf{x}_i^{\text{cmd}}\|}{x_i^{\max}}\right) \frac{\mathbf{x}_i^{\text{cmd}}}{\|\mathbf{x}_i^{\text{cmd}}\|},  
\end{equation}  
where \(\mathbf{x}_i^{\text{cmd}} \in \{\mathbf{v}_i^{\text{cmd}}, \mathbf{u}_i^{\text{cmd}}\}\) is the commanded (unsaturated) input, \(\mathbf{x}_i\in \{\mathbf{v}_i, \mathbf{u}_i\}\) is the applied (saturated) input and \(x_i^{\max} \in \{v_i^{\max}, u_i^{\max}\}\) is the corresponding maximum  magnitude (with the convention \(\mathbf{0}/\|\mathbf{0}\| = \mathbf{0}\)).

The model balances cohesion-separation via interaction weight $\psi$ and alignment via $\phi$. For cohesion-separation, $\psi<0$ (repulsive) when $\|\mathbf{p}_j - \mathbf{p}_i\| < \delta_i |\mathcal{N}_i|$ ensures short-range repulsion; $\psi>0$ (attractive) when $\|\mathbf{p}_j - \mathbf{p}_i\| > \delta_i |\mathcal{N}_i|$ promotes long-range attraction; and $\psi=0$ at $\|\mathbf{p}_j - \mathbf{p}_i\| = \delta_i |\mathcal{N}_i|$ marks equilibrium spacing. Smaller $\delta_i$ yields denser formations, larger $\delta_i$ looser ones, with built-in collision avoidance. The alignment weight $\phi>0$ drives velocity consensus.

The purely reactive flocking model in~\eqref{eq:flock-model} uses instantaneous neighbor states, whereas anticipating neighbor motions improves coordination in both natural and engineered systems. We therefore propose the FDA framework, in which agents integrate short-horizon velocity predictions of neighbors—obtained via communication or estimation—directly into the interaction rules, thereby enhancing collective dynamics. The approach is detailed in the next section.

\section{FDA Flocking with Velocity Prediction}\label{sec:fda_flocking}

Agent \(i\) has access to the current states of neighbors \(j \in \mathcal{N}_i\): \(\mathbf{p}_j\), \(\mathbf{v}_j\), and \(\mathbf{u}_j\) (via estimation or communication). It predicts the velocity as
\begin{equation}\label{eq:dynamics-predictive}
\mathbf{v}_j^{\text{pred}} = \mathbf{v}_j + t_i^{\text{ph}} \mathbf{u}_j,
\end{equation}
where $t_i^{\text{ph}} > 0$ is the prediction horizon. The result is saturated by~\eqref{eq:saturation}.

The FDA flocking model, incorporating velocity predictions into agent interactions, is given by
\begin{align} \label{eq:FDA-Flocking}
\mathbf{u}_i =&\sum_{j \in \mathcal{N}_i} \psi(\|\mathbf{p}_j - \mathbf{p}_i\|)(\mathbf{p}_j - \mathbf{p}_i) \notag\\&+(1-\theta_i)\phi_i \sum_{j \in \mathcal{N}_i}  (\mathbf{v}_j - \mathbf{v}_i)+\theta_i\phi_i \sum_{j \in \mathcal{N}_i}  (\mathbf{v}_j^{\text{pred}} - \mathbf{v}_i),
\end{align}
where \(\theta_i \in [0,1]\) blends current (\(\theta_i=0\), reactive) and predicted (\(\theta_i=1\), fully anticipatory) alignment terms.

The predictive velocity term \(\mathbf{v}_j^{\text{pred}}\) in \eqref{eq:FDA-Flocking} serves as a mathematical analogue to anticipatory signals, encoding intended motion directions akin to birds' postural cues or drones' rotor tilts. By blending reactive and predictive alignment via \(\theta_i\), the model mimics natural collectives and supports engineered swarms, where such cues facilitate smoother trajectory corrections and emergent coordination without relying solely on current states. 

In quadrotor swarms, the quantities used to form \(\mathbf{v}_j^{\text{pred}}\) correspond to directly sensed or communicated signals (e.g., acceleration reconstructed from onboard tilt (pitch/roll) measurements that determine thrust direction together with thrust magnitude inferred from propeller speeds), rather than relying on explicit state observers or filter-based estimation.

\subsection{Alignment Convergence under Prediction}

To gain insight into how velocity prediction affects alignment, consider the idealized case of steady-state flocking under the following assumptions: (i) fixed, connected interaction graph, (ii) no control saturation, (iii) uniform prediction horizon $t^{\mathrm{ph}}$ and uniform blending parameter $\theta$ across agents, (iv) cohesion–separation forces at equilibrium (negligible contribution to velocity dynamics), and (v) absence of delay and noise. Under near-alignment conditions, where cohesion–separation terms vanish at equilibrium and saturation effects are inactive, the FDA control law \eqref{eq:FDA-Flocking} admits the approximation
$$
\dot{\mathbf{v}}_i \approx \phi \sum_{j \in \mathcal{N}_i} \Big[ (1-\theta)(\mathbf{v}_j - \mathbf{v}_i) + \theta (\mathbf{v}_j + t^{\mathrm{ph}} \dot{\mathbf{v}}_j - \mathbf{v}_i) \Big]=\phi \sum_{j \in \mathcal{N}_i} (\mathbf{v}_j - \mathbf{v}_i) + \theta \phi t^{\mathrm{ph}} \sum_{j \in \mathcal{N}_i} \dot{\mathbf{v}}_j,
$$
where $\dot{\mathbf{v}}_j \approx  \mathbf{u}_j$. In vector form, with $\mathbf{v} = [\mathbf{v}_1^\top, \dots, \mathbf{v}_n^\top]^\top$, the Laplacian $\mathbf{L}$, and adjacency matrix $\mathbf{A}$, the collective velocity dynamics, after rearrangements, become
$$
\dot{\mathbf{v}} \approx  -\phi \big( \mathbf{I} + \theta \phi t^{\mathrm{ph}} (\mathbf{A} \otimes \mathbf{I}_m) \big)^{-1} (\mathbf{L} \otimes \mathbf{I}_m) \mathbf{v}.
$$

For a connected graph, $\mathbf{L}$ has one zero eigenvalue (consensus subspace) and positive eigenvalues elsewhere. The matrix $\mathbf{I} + \theta \phi t^{\mathrm{ph}} \mathbf{A} \otimes \mathbf{I}_m$ is invertible for typical (small) values of $\theta \phi t^{\mathrm{ph}}$. The resulting system matrix is negative semi-definite on the disagreement subspace, so velocity differences $\mathbf{v}_i - \mathbf{v}_j$ decay exponentially to zero—i.e., asymptotic velocity consensus (alignment) is preserved.

The predictive term effectively preconditions the consensus dynamics, accelerating convergence for moderate $t^{\mathrm{ph}}$ and $\theta$. However, excessively large prediction horizons with large blending weights can push eigenvalues toward instability, imposing a practical trade-off between anticipatory benefit and robustness. This analysis provides theoretical support for the observed faster and more robust alignment in FDA flocking (in Section~\ref{sec:conclusion}) while highlighting the limits of naive long-horizon prediction.

\subsection{Incorporation of Delay and Noise}
\label{subsec:delay-noise}

In realistic settings, agents experience communication/measurement delays and sensor noise. For agent $i$, the delayed states of neighbor $j$ are $\mathbf{p}_j^{\tau_i}$, $\mathbf{v}_j^{\tau_i}$, and $\mathbf{u}_j^{\tau_i}$ (shifted by agent-specific delay $\tau_i \geq 0$). Delayed states are used directly; FDA predicts via open-loop extrapolation of delayed velocity using delayed acceleration. The perceived states are further corrupted by zero-mean Gaussian noise, $\tilde{\mathbf{p}}_j^{\tau_i} = \mathbf{p}_j^{\tau_i} + \mathbf{n}_{p,ij},\tilde{\mathbf{v}}_j^{\tau_i} = \mathbf{v}_j^{\tau_i} + \mathbf{n}_{v,ij},
\tilde{\mathbf{u}}_j^{\tau_i} = \mathbf{u}_j^{\tau_i} + \mathbf{n}_{u,ij}$, with $\mathbf{n}_{p,ij} \sim \mathcal{N}(\mathbf{0}, \sigma_p^2 \mathbf{I})$, $\mathbf{n}_{v,ij} \sim \mathcal{N}(\mathbf{0}, \sigma_v^2 \mathbf{I})$, $\mathbf{n}_{u,ij} \sim \mathcal{N}(\mathbf{0}, \sigma_u^2 \mathbf{I})$, and \(\sigma_p\), \(\sigma_v\), and \(\sigma_u\) represent the respective noise intensities. The resulting FDA control law under delay and noise becomes
\begin{equation}
\begin{aligned}
\mathbf{u}_i
&= \sum_{j \in \mathcal{N}_i}
\psi\big(\|\tilde{\mathbf{p}}_j^{\tau_i} - \mathbf{p}_i\|\big)
\big(\tilde{\mathbf{p}}_j^{\tau_i} - \mathbf{p}_i\big) \\
&\quad + (1-\theta_i)\phi_i \sum_{j \in \mathcal{N}_i}
\big(\tilde{\mathbf{v}}_j^{\tau_i} - \mathbf{v}_i\big)
+ \theta_i \phi_i \sum_{j \in \mathcal{N}_i}
\big(\tilde{\mathbf{v}}_j^{\mathrm{pred}} - \mathbf{v}_i\big),
\end{aligned}
\label{eq:fda_delay_noise}
\end{equation}
where $\tilde{\mathbf{v}}_j^{\mathrm{pred}} = \tilde{\mathbf{v}}_j^{\tau_i} + t_i^{\mathrm{ph}} \tilde{\mathbf{u}}_j^{\tau_i}$ (and saturated as before).

By extrapolating from delayed and noisy data, the predictive term reduces the adverse impact of communication lag—by approximating neighbors’ near-future states—and attenuates high-frequency stochastic perturbations through short-horizon forward projection. This bio-inspired mechanism, motivated by anticipatory cues observed in natural flocks, enhances robustness to timing errors and measurement uncertainty without relying on explicit delay estimators or noise filtering. 

\section{Simulation Results} \label{sec:simulation_results}

To validate the FDA flocking framework—motivated by predictive postural cues observed in birds and drones—simulations compare it against the reactive model. Simulations use $n=10$ agents in 3D space over $T=25\,\text{s}$ with time step $dt=0.02\,\text{s}$. Each agent employs a prediction horizon \( t_i^{\mathrm{ph}} = 1\,\mathrm{s} \) and experiences a communication delay \( \tau_i = 0.4\,\mathrm{s} \). The prediction horizon is chosen in line with common practice in short-horizon predictive control for fast dynamical systems; smaller horizons reduce the FDA model toward the reactive baseline, while larger horizons amplify prediction errors and may destabilize collective motion. The delay duration reflects typical sensing and communication latencies reported for multirotor drone swarms. Agent dynamics are bounded by \( v_i^{\max} = 4\,\mathrm{m/s} \) and \( u_i^{\max} = 8\,\mathrm{m/s}^2 \), representative of common multirotor platforms; these bounds primarily affect transient timing rather than qualitative flocking behavior. Initial positions are sampled uniformly from \( [0,10)^m \), and initial velocities are drawn i.i.d. from \( \mathcal{N}(\mathbf{0},1) \). The interaction radius \( r_i = 7.5\,\mathrm{m} \) and spatial offset \( \delta_i = 1 \) are selected to yield an average neighborhood size of approximately 4–7 agents, consistent with biological and robotic swarm studies. The blending parameter \( \theta_i = 0.8 \) is chosen to emphasize the predictive component; smaller values smoothly recover reactive flocking behavior.

\begin{figure}[t]
\centering
\includegraphics[width=1\textwidth]{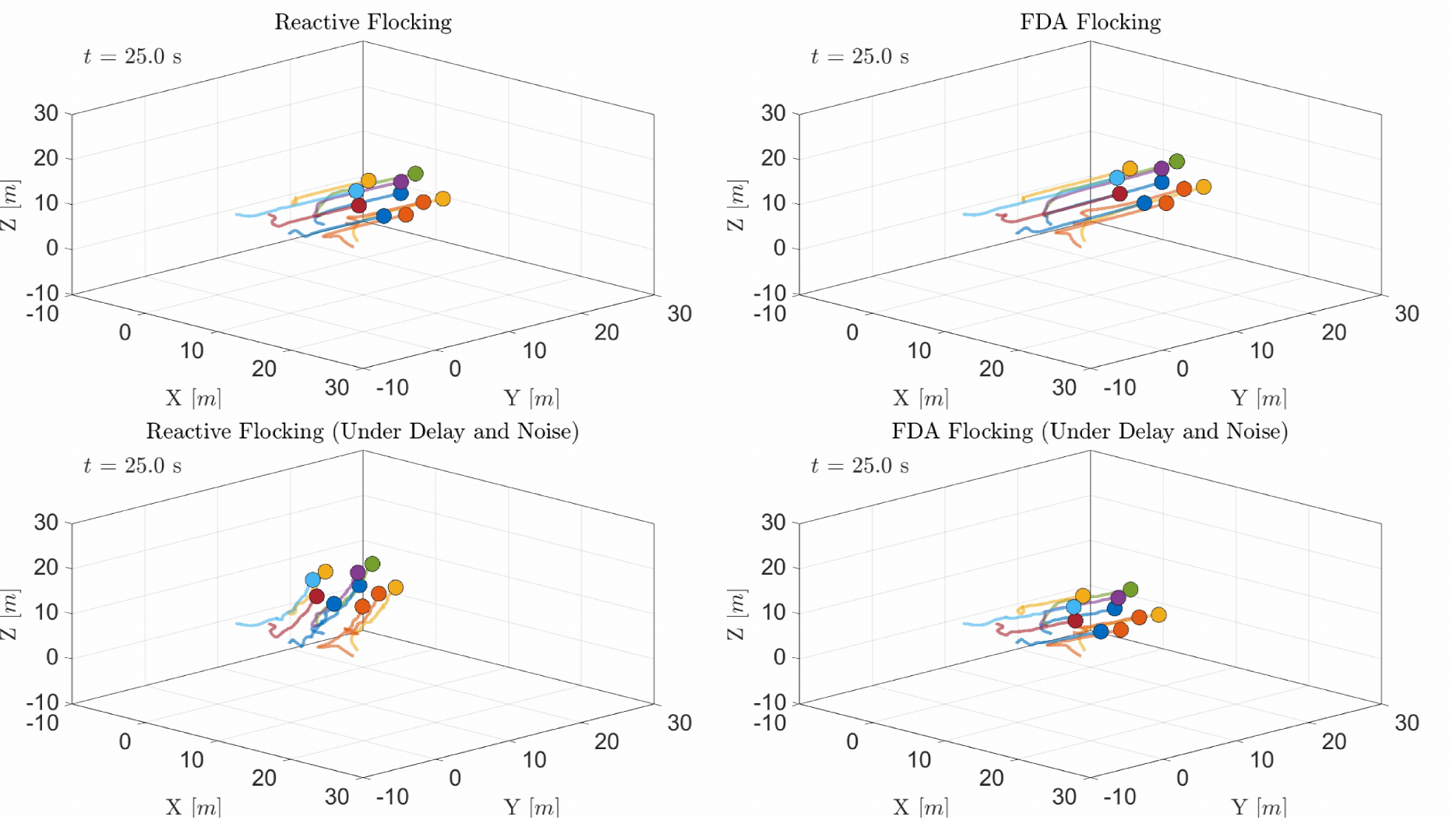}
\caption{3D agent trajectories over $t \in [0, 25]\,\text{s}$: reactive (left) and FDA (right); nominal conditions (top) and with delay and noise (bottom).}
\label{fig:traj}
\end{figure}

\begin{figure}
    \centering
    \begin{minipage}{0.32\textwidth}
        \centering
        \includegraphics[width=\linewidth]{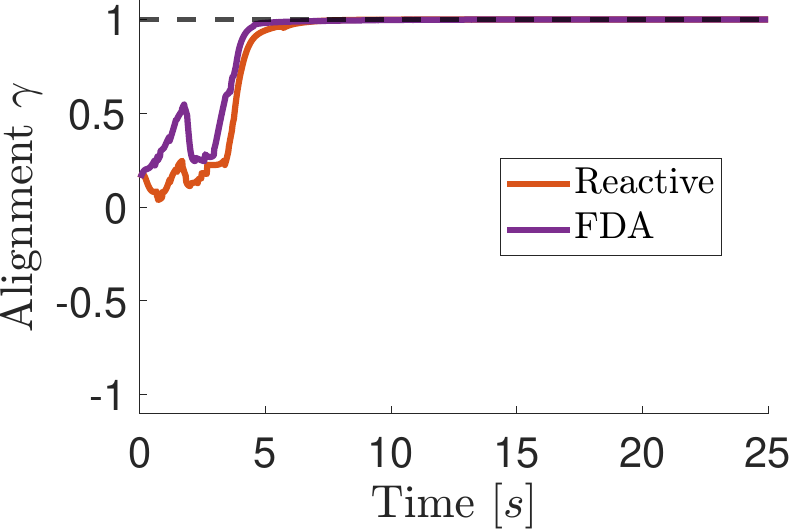}
        {\small (a)}
    \end{minipage}
        \hfill
    \begin{minipage}{0.32\textwidth}
        \centering
        \includegraphics[width=\linewidth]{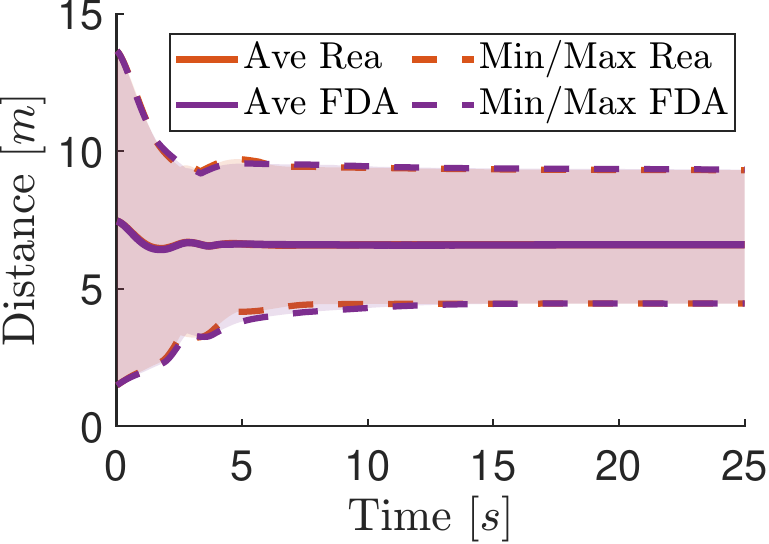}
        {\small (b)}
    \end{minipage}
    \hfill
    \begin{minipage}{0.32\textwidth}
        \centering
        \includegraphics[width=\linewidth]{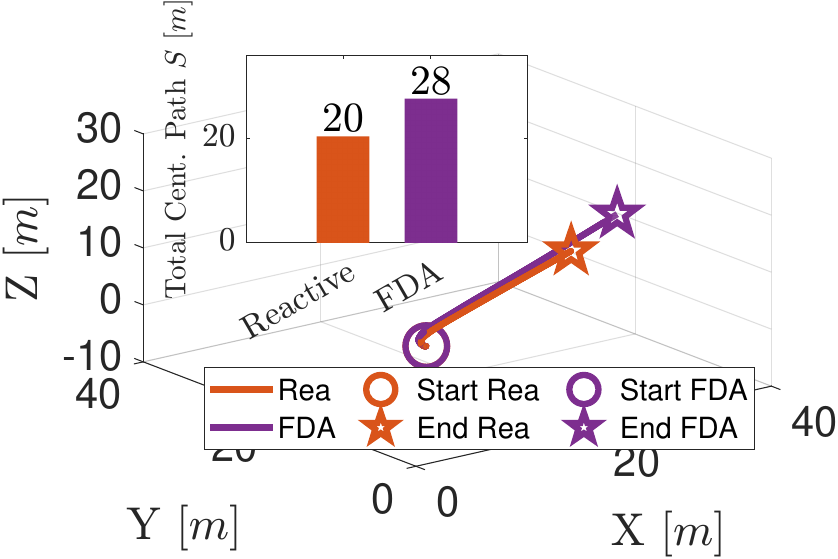}
        {\small (c)}
    \end{minipage}
    \begin{minipage}{0.32\textwidth}
        \centering
        \includegraphics[width=\linewidth]{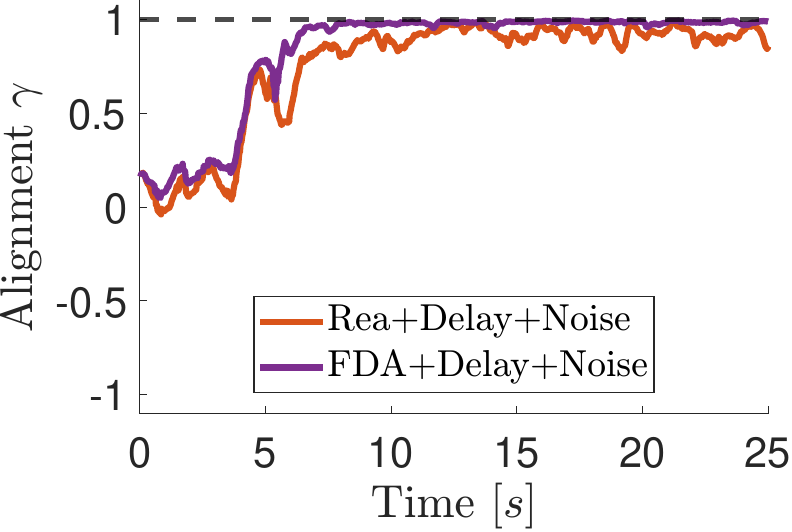}
        {\small (d)}
    \end{minipage}
    \hfill
    \begin{minipage}{0.32\textwidth}
        \centering
        \includegraphics[width=\linewidth]{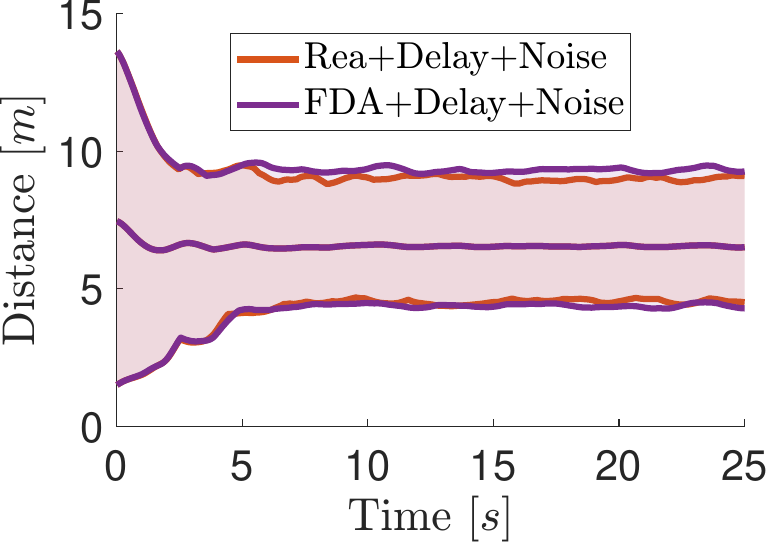}
        {\small (e)}
    \end{minipage}
    \hfill
    \begin{minipage}{0.32\textwidth}
        \centering
        \includegraphics[width=\linewidth]{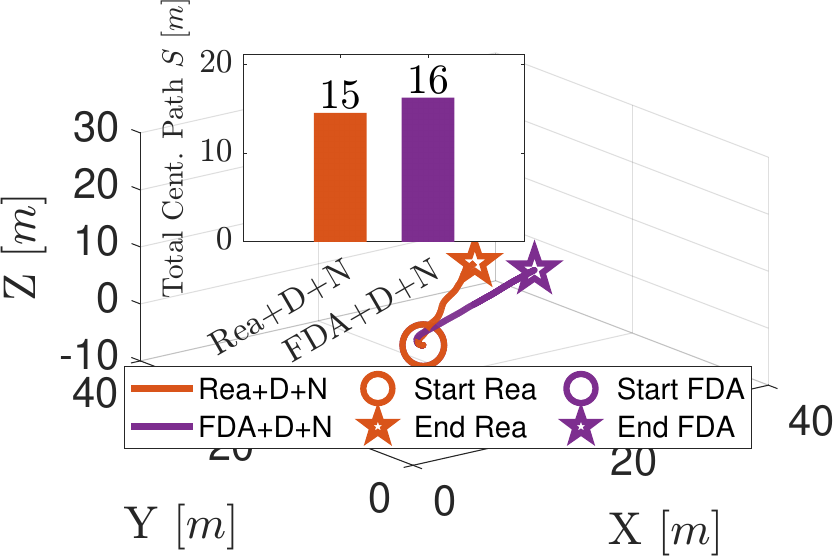}
        {\small (f)}
    \end{minipage}
\caption{Metrics for reactive and FDA flocking: nominal (top) vs. delay + noise (bottom). (a,d)~alignment $  \gamma  $; (b,e)~inter-agent distances; (c,f)~centroid paths with length $  S  $.}
    \label{fig:flock-hist}
\end{figure}

Flocking performance is compared between the reactive and the FDA variant using the following metrics:
(i) \textit{directional alignment} ($\gamma \in [-1,1]$): average pairwise cosine similarity of velocities between each agent and its neighbors (averaged across the flock), where values near 1 indicate strong velocity coherence~\cite{jond2025position}; (ii) \textit{inter-agent distances}: minimum, mean, and maximum pairwise distances between agents; and (iii) \textit{flock centroid path length} ($S$): total distance traveled by the flock centroid over the simulation. Both reactive and FDA flocking achieve strong alignment and cohesion under nominal conditions (top row, Fig.~\ref{fig:traj}), from the same random initialization. The corresponding time histories in Fig.~\ref{fig:flock-hist} (top row) compare (a) alignment $\gamma$, (b) inter-agent distances, and (c) centroid trajectories with $S$. Subfigures (a)--(c) highlight the FDA's efficiency. In (a), FDA reaches peak $\gamma$ earlier with consistently higher values, aiding cohesion via prediction. As shown in (b), both models yield comparable distance profiles. Centroids in (c) align, with FDA's $S$ being 40\% larger. Multiple runs confirm these trends.

A second scenario incorporates delays and time-varying zero-mean Gaussian noise in position, velocity, and acceleration data to assess FDA robustness (Fig.~\ref{fig:traj}, bottom row). Noise standard deviations vary asynchronously to mimic environmental fluctuations, \(
\sigma_p = 0.5 + 0.10 \sin(5t)\,\text{m},\sigma_v = 0.2 + 0.05 \sin(5 t + \pi/4)\,\text{m/s},\sigma_u = 0.1 + 0.02 \sin(5t + \pi/2)\,\text{m/s}^2\). The time histories in Fig.~\ref{fig:flock-hist} (bottom row) show that FDA exhibits only minor performance degradation under these adverse perturbations. Although delays and noise slow alignment, the predictive structure of the FDA enables the collective motion to maintain near-nominal alignment and centroid stability. In contrast, the reactive model does not consistently achieve full alignment and exhibits lower efficiency, with centroid trajectories deviating substantially from nominal behavior. For both models, inter-agent distance profiles remain largely comparable to nominal conditions. These results demonstrate the robustness of the FDA under sensing and communication constraints.

\section{Conclusion}
\label{sec:conclusion}

We introduced Future Direction-Aware (FDA) flocking, a bio-inspired anticipatory extension of reactive flocking that enables proactive coordination. FDA blends reactive alignment with a predictive term via a tunable blending parameter, balancing immediate feedback with forward-looking adjustments. Simulations confirm FDA’s advantages over the reactive baseline, including faster and higher alignment and substantially less performance degradation under adverse perturbations (delay and Gaussian noise). FDA enhances the robustness of collective motion under realistic sensing and communication constraints, supporting its potential for scalable deployment in aerial robotic swarms. Limitations of the current model include uniform averaging of velocity predictions, which may overlook biologically observed differential weighting of stronger directional cues. Future work will explore weighted prediction schemes and empirical validation.

\subsection*{Acknowledgments}
This work was funded by the Czech Science Foundation (GAČR) under research project no. 23-07517S and the European Union under the project Robotics and Advanced Industrial Production (reg. no. CZ.02.01.01\slash00\slash22\_008\slash0004590).

\bibliographystyle{splncs04}
\bibliography{references}

\end{document}